\documentclass[letterpaper]{article} 
\usepackage{aaai2026}
\usepackage{times}  
\usepackage{helvet}  
\usepackage{courier}  
\usepackage[hyphens]{url}  
\usepackage{graphicx} 
\usepackage{float}
\usepackage{natbib}  
\usepackage{caption} 
\usepackage{algorithm}
\usepackage{algorithmic}
\usepackage{amsmath} 
\usepackage{graphicx,subcaption}
\usepackage{amssymb}
\usepackage{newfloat}
\usepackage{listings}
\DeclareCaptionStyle{ruled}{labelfont=normalfont,labelsep=colon,strut=off} 
\floatstyle{ruled}
\newfloat{listing}{tb}{lst}{}
\floatname{listing}{Listing}
\title{LTM: Large-scale Terrain Model for Landscapes}
\author{
    Xiao Fu, Yue Hu, Meida Chen, Peter Anthony Beerel, Barath Raghavan \\
}
\affiliations{
    University of Southern California\\

}

\usepackage{bibentry}

\begin{document}

\maketitle

\begin{abstract}

Accurate 3D terrain maps are essential for emergency response when assessing wildfire hazards. However, wildfire-prone regions often span vast areas where conventional reconstruction methods underperform. Airborne LiDAR systems provide high-resolution terrain data, but they are expensive and infrequently updated. Image-based methods offer a lower-cost alternative, but struggle due to sparse visual features and limited image overlap. We propose a multi-modal reconstruction framework leveraging outdated Digital Elevation Models (DEMs) as geometric priors for image-based 3D reconstruction. Our key innovation is physics-based pixel-pixel alignment between images and DEM data, dramatically reducing computational complexity by eliminating expensive feature matching procedures. To validate our approach, we developed a large-terrain simulator based on a real wildfire-prone area, generating realistic images enabling a comprehensive evaluation. Given posed images and legacy DEMs, our method produces high-fidelity depth maps while maintaining real-time performance. We find significant improvements in reconstruction accuracy and computational efficiency over existing techniques, offering a scalable solution for wildfire response.
\end{abstract}

\section{Introduction}

\label{sec:intro}
Environmental monitoring has yet to fully leverage the vast network of cameras already deployed in wildfire-prone and other environmentally vulnerable regions. These camera systems capture rich spatio-temporal imagery that offers valuable insights into dynamic natural landscapes. Such imagery can help identify both gradual ecological changes, such as vegetation growth, and sudden environmental events, including avalanches~\cite{barbolini2011avalanche}, floods~\cite{mudashiru2021flood}, and wildfires~\cite{fu2024fireloc}. 

Accurate mapping of such landscapes not only enhances disaster preparedness but also aids automated semantic analysis, including assessments of disaster intensity. In particular, fuel maps, which provide vegetation-related spatial semantics, are widely used in conjunction with terrain models to support wildfire propagation simulators such as FARSITE~\cite{finney1998farsite} and FlamMap~\cite{finney2006FLAMMAP}. In addition, 3D reconstruction techniques generate detailed representations of natural terrain, which serve as the spatial foundation on which fuel maps are overlaid, enabling the extraction of timely and accurate insights that improve early warning systems and support more informed and responsive disaster management.

\begin{figure}[t]
  \centering
   \includegraphics[width=1\linewidth]{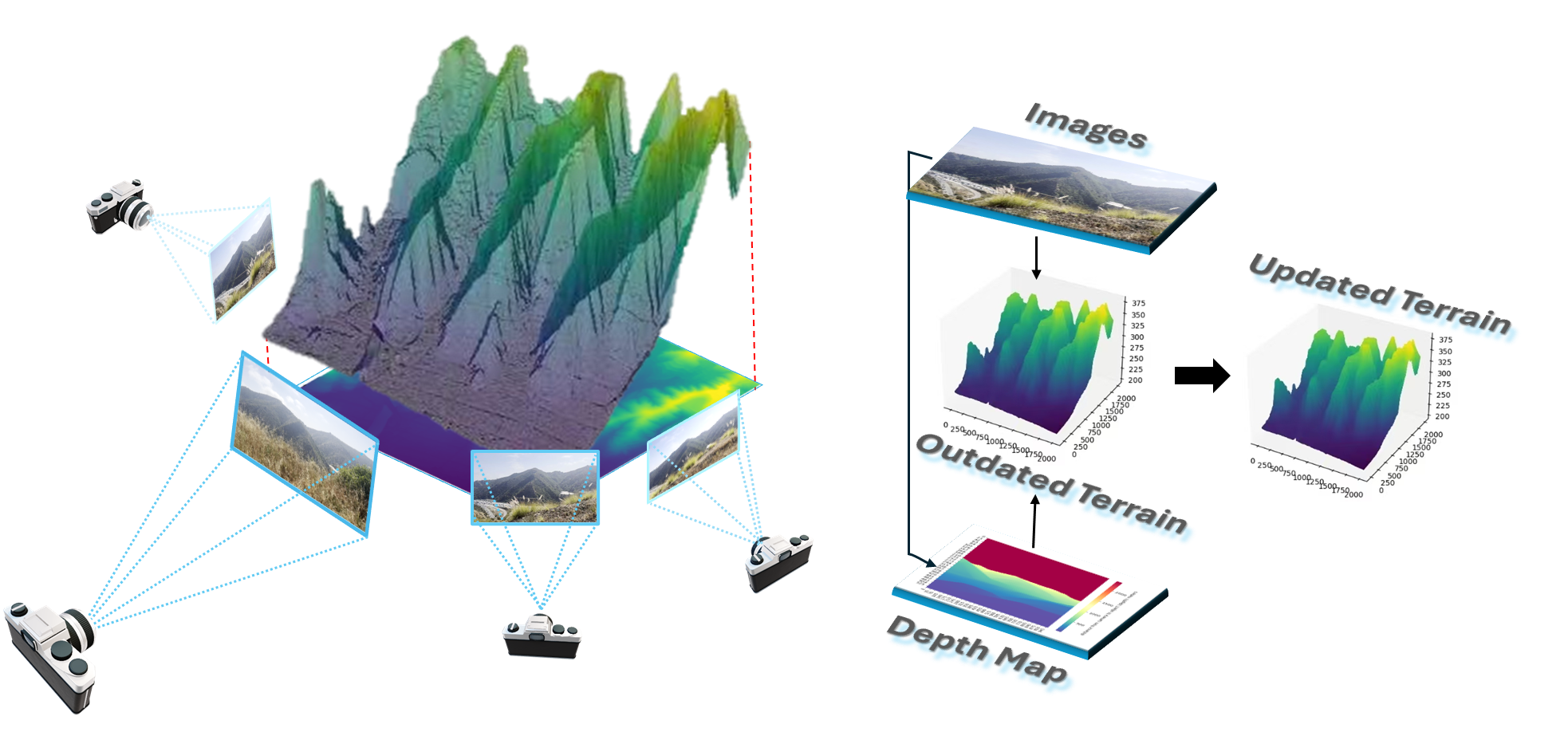}

   \caption{A novel framework for large-scale dynamic terrain updates in end-to-end 3D semantic mapping.}
   \label{fig:teas}
\end{figure}

Landscape, terrain, and vegetation are traditionally captured using remote sensing methods using spacecraft~\cite{hirano2003spacemapping} or aircraft-based platforms~\cite{dobrowski2008mapping, rodriguez2002matching}. However, the resulting 3D models are typically updated only on an annual basis, if that, due to the vast areas that must be scanned~\cite{gorelick2017google,krishnan2011opentopography, kervyn2007mapping}. This infrequent update cycle is insufficient for effective disaster mitigation, as significant changes in vegetation and surface conditions can occur over seasonal or even monthly timescales~\cite{yang2011srtm}. In contrast, ground-based camera networks offer higher temporal resolution, but lack the top-down, wide-area coverage that is critical for comprehensive landscape monitoring.

While prior ground image-based 3D reconstruction methods have shown strong performance in urban environments, their effectiveness in wildfire-prone natural landscapes remains limited~\cite{iglhaut2019structure}. These approaches typically focus on estimating camera pose and scene structure from image sequences, enabling spatial inference of objects and surfaces. However, their applicability to vegetated environments is constrained by several factors. Vegetation often lacks color contrast and is constantly changing structurally due to wind and seasonal growth~\cite{nguyen2015structured, sidle2017strengths, schmidt2025rover}. At the same time, the large spatial extent covered by each ground-level camera often leads to sparse image overlap, reducing feature correspondence and increasing reconstruction errors~\cite{deng2025gigaslam}. In addition, hand-crafted feature descriptors such as SIFT~\cite{lowe2004distinctive} and ORB~\cite{rublee2011orb} struggle to detect stable, repeatable keypoints in vegetation-dominated scenes due to low texture, occlusions, and dynamic patterns~\cite{qadri2021semantic}. Ground-level landscape imagery is often spatially sparse and temporally asynchronous, further complicating model-based reconstruction in large-scale terrain. Moreover, unlike urban environments---which exhibit fine-grained, centimeter-level geometry and rapid second-level changes---natural landscapes require meter-level spatial resolution~\cite{hodgson2003evaluation} and evolve over weekly to monthly timescales, demanding fundamentally different reconstruction assumptions.

Existing image-based methods use various 3D representations to achieve more accurate scene rendering and mapping, such as neural radiance fields (NeRF) to enhance the accuracy of reconstruction~\cite{mildenhall2021nerf}. Inspired by this approach, we adopt a more suitable representation, DEM, for landscape applications instead of relying on voxel-based methods. However, existing image-based methods are predominantly designed for urban environments because of the scarcity in available large-scale vegetated datasets. Even when applying state-of-the-art large-scale NeRF 3D reconstruction techniques~\cite{turki2022mega}, performance degrades significantly in non-urban natural landscape scenes~\cite{deng2025gigaslam, lu2023large, hermann2024leveraging}. For example, satellite-based NeRF approaches such as CityNeRF~\cite{turki2022mega} incorporate digital surface models (DSM) into their evaluation pipeline but remain optimized for urban settings. We argue that existing state-of-the-art 3D representations are insufficient for accurate and efficient mapping and reconstruction in landscape environments. Therefore, we propose an improved method for mapping in landscape spatial monitoring that uses secondary scene information, particularly depth maps generated from ground-based RGB images.

We present a cross-modality landscape mapping approach that accurately detects changes in scenes while updating scene-captured 3D models as presented in Figure~\ref{fig:teas}. First, we adopt a vertical historical large-scale baseline model using a DEM in raster format. Second, we leverage existing learning-based monocular depth estimation tools to generate horizontal depth information from ground-based imagery. Third, by combining vertical and horizontal spatial information with automated identification of changing sections, we can accurately update the landscape by achieving consensus across multiple images. Through the identification of these shifting landscape features, we can dynamically update the baseline raster models to maintain current and accurate environmental representations.\\



\noindent\textbf{Contribution.} 
\begin{itemize}
\item \textbf{Monocular depth mapping.} We develop a self-supervised monocular depth estimation system specifically designed for large-scale vegetated outdoor environments, addressing the challenges of feature extraction in wildfire-prone landscapes.

\item  \textbf{Outdoor cross-modal dataset.} We compile and evaluate existing cross-modality datasets and graphic-based simulators focused on wildfire-prone environments, providing comprehensive benchmarks for localization and mapping accuracy in large-scale wildfire-prone settings.

\item \textbf{3D representation and segmentation for large-scale outdoor landscapes.} We introduce the strategic use of Digital Elevation Models (DEMs) as 2D arrays that effectively capture 3D environmental information. Our approach successfully achieves robust 2D-3D alignment and improves existing rendering and mapping frameworks. This method specifically provides alignment of each image pixel to the DEM raster pixel.

\end{itemize}

\section{Related Work}
\subsection{Spatial Representation and Reconstruction}
Traditional 3D representations, including point clouds~\cite{rusu20113dpcl}, meshes, and voxel grids~\cite{wu20153dvoxel}, excel in urban reconstruction with high fidelity and flexible visualization. Combined with feature descriptors like ORB~\cite{rublee2011orb} and SIFT~\cite{lowe2004distinctive}, these representations increasingly adopt neural and statistical methods to enhance accuracy and visual quality.
Traditional 3D reconstruction methods rely heavily on feature descriptors through approaches such as Multi-view Stereo (MVS)~\cite{schoenberger2016mvs} and Simultaneous Localization and Mapping (SLAM)~\cite{mur2015orb}. Recent advances, particularly Neural Radiance Fields (NeRF)~\cite{mildenhall2021nerf} and 3D Gaussian Splatting methods like PixelSplat~\cite{charatan2024pixelsplat}, have shown promise but face significant challenges in non-urban scenarios~\cite{mall2023change}.

\subsection{2D-3D Alignment and Multi-modal Improvement}
Feature matching and 2D-3D alignment is essential for 3D reconstruction. Prior work has explored cross-modality registration between 2D images and 3D representations, including implicit (e.g., NeRF) and explicit (e.g., LiDAR point clouds)~\cite{zhou2023differentiable, wu2023virtual, li2022deepfusion, chen2024map, bhunia2024looking, cho2025zero}. Line correspondences offer an alternative when visual features are sparse~\cite{yu2020monocular}. Key challenges remain, including runtime-storage trade-offs~\cite{brachmann2023accelerated} and large search spaces~\cite{li2021deepi2p}. While neural feature matchers such as LoFTR~\cite{sun2021loftr}, SuperGlue~\cite{sarlin2020superglue}, and others~\cite{panek2022meshloc, cheng2025bridge} achieve robust results, they remain computationally intensive and less effective in large-scale, vegetated settings.



\subsubsection{DEM-supported vision tasks}
 Prior work~\cite{liu2022random} has described the advantages of DEMs in large-scale terrain representation, particularly their efficiency and storage benefits. Early work has explored using DEM models for outdoor geo-localization~\cite{baatz2012large}, which addresses the difficulty of feature recognition by using skyline-shaped mountain ridges as matching points. 

\subsection{3D Reconstruction and Semantics}
\subsubsection{Sparse images for vast area}
Large-scale environmental monitoring typically relies on satellite-based systems~\cite{mall2023change,turki2022mega} for broad coverage; however, their limited temporal and spatial resolution motivates the development of complementary image-based reconstruction methods. Recent neural approaches, including DUST3R~\cite{dust3r_cvpr24}, POW3R~\cite{jang2025pow3r}, and MAST3R~\cite{mast3r_eccv24}, demonstrate effective performance in sparse image scenarios by integrating image features and geometric priors for efficient 3D reconstruction.

\subsubsection{Depth estimation}

Early depth estimation approaches relied on multi-view stereo~\cite{li2018megadepth, godard2019digging}, but demonstrated poor performance in outdoor vegetated environments. Recent advances in monocular depth estimation have achieved impressive results using only RGB images as input, with state-of-the-art methods including UniDepth~\cite{piccinelli2024unidepth, piccinelli2025unidepthv2}, Depth Anything~\cite{depth_anything_v1, depth_anything_v2}, DepthPro~\cite{bochkovskii2024depth}, Metric3D V2~\cite{hu2024metric3d}, and Marigold~\cite{ke2025marigold}. These models broaden the applicability of depth estimation across diverse settings while maintaining efficiency and scalability.




\subsubsection{3D semantic segmentation}
Recent advances in end-to-end 3D semantic segmentation have demonstrated significant benefits for scene understanding by integrating volumetric reconstruction with semantic labeling. Methods like LSM~\cite{fan2024largespatialmodelendtoend} show how 3D segmentation networks generate spatially-coherent semantic representations. These end-to-end 3D segmentation techniques that extend beyond open-vocabulary 2D methods like LSeg~\cite{li2022languagedriven} offer particular promise for large-scale landscape semantic analysis.


\section{Preliminaries}

\subsection{Digital Elevation Models}
DEMs~\cite{mukherjee2013evaluation} are typically represented using raster data structures. While this sacrifices geometric detail compared to mesh or point cloud formats, it offers significant computational advantages for mega-scale applications spanning cities or wildlands. For landscapes exceeding 1 square km in scale, raster-based DEMs as a 3D representation provide an ideal balance between representational fidelity and computational efficiency. This efficiency becomes particularly valuable when integrating outdated landscape information with real-time imagery for image-based landscape mapping, enabling scalable reconstruction pipelines that can process vast geographic areas while maintaining reasonable computational requirements. OpenTopography~\cite{opentopowebsite, krishnan2011opentopography} provides an annual DEM update based on the results of the LiDAR survey. 

\begin{figure}[!t]
\centering
\begin{subfigure}[t]{.49\linewidth}
    \centering
\includegraphics[width=\linewidth]{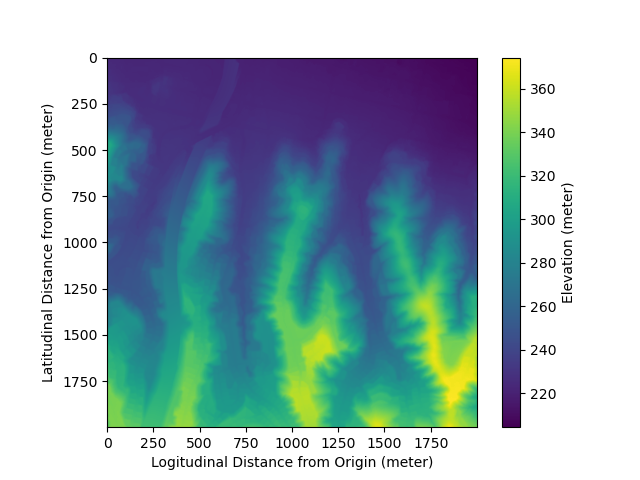}
\subcaption{Storage}
\end{subfigure}
\begin{subfigure}[t]{.49\linewidth}
    \centering
    \includegraphics[width=\linewidth]{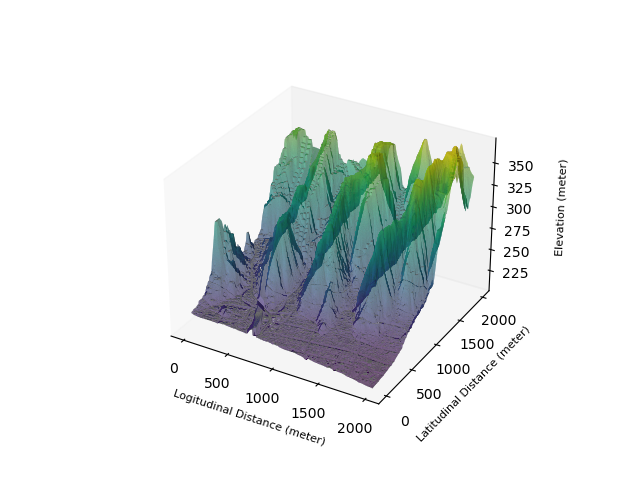}
    \subcaption{Visualization}
\end{subfigure}

\caption{DEM efficiently represents large-scale 3D terrain through per-pixel elevation storage. This 16 $km^2$ area shows detailed ravines and ridges.}
\label{fig:DEMViz}
\end{figure}

As presented in Figure~\ref {fig:DEMViz}, the 2D raster on the right presents a $4000\,\mathrm{m} \times 4000\,\mathrm{m}$ area where each pixel of $1\,\mathrm{m} \times 1\,\mathrm{m}$ is associated with the elevation of such a location. The left presents the same terrain projected into a 3D space, as you can see the gully and ridges in the mountainous area. The area shown covers all that was impacted by the 2019 Getty Fire.

\begin{table*}[t!]
\centering
\resizebox{2.1\columnwidth}{!}{
\begin{tabular}{c|cccc}
\hline
Method                                                    & Task                                 & 3D Representation                    & Mapping Complexity (W/ Poses)              & Mapping Complexity (W/o Poses)           \\ \hline
NeRF (Neural Radiance Fields)                             & Novel View Synthesis                 & Volume (Implicit)                    & O(NvMP)                                       & O(Nf\textasciicircum{}2MP)                    \\
Gaussian Splatting                                        & Novel View Synthesis                 & Point Cloud (Implicit)              & O(NpMP)                                       & O(Nf\textasciicircum{}2MP)                    \\
Gaussian Splatting + Process                     & Uncertainty-aware Rendering          & Point Cloud (Implicit) & O(Np\textasciicircum{}3+NpMP)                 & O(Nf\textasciicircum{}2MP)                    \\
Multi-view Stereo (COLMAP)                            & Dense 3D Reconstruction     & Point Cloud/Mesh                     & O(NfMP)                                       & O(Nf\textasciicircum{}2MP)                    \\
SLAM              & Real-time Mapping & Point Cloud / Mesh                   & O(NfMP)                                       & O(Nf\textasciicircum{}2MP)                    \\
Radiant Foam                                              & Real-time Ray Tracing                & Volume (Sparse Voxels)               & O(NvMP)                                       & O(Nf\textasciicircum{}2MP)                    \\
Monocular Depth Estimation                                & Single-Image Depth Prediction        & Depth Map                            & O(NfMP)                                       & O(Nf\textasciicircum{}2MP)                    \\
2D-3D Line Correspondence & Pose Estimation from 3D Map          & 3D Lines (LiDAR Map)                 & O(NfMP)                                       & O(Nf\textasciicircum{}2MP)                    \\
SuperGlue (Feature Matching)                              & Feature Matching Across Images       & N/A (Keypoints)                      & O(Nf\textasciicircum{}2MP\textasciicircum{}2) & O(Nf\textasciicircum{}2MP\textasciicircum{}2) \\
DEM-Supported Methods                                     & Large-Scale Terrain Mapping          & Elevation Map                        & O(NvMP)                                       & O(Nf\textasciicircum{}2MP)                    \\ \hline
\end{tabular}}\caption{Compared to other 3D representations, DEM for landscape in 3D modeling can largely reduce the computation cost for cross-modality tasks for 3D reconstruction in creating 2D-3D correspondence.}
\label{table:compcost3D}
\end{table*}
\subsection{Fuel Map}
Large-scale terrain semantic segmentation provides essential fuel information for fire behavior analysis and risk assessment. LANDFIRE~\cite{rollins2009landfire} provides fuel maps with $30\,\mathrm{m} \times 30\,\mathrm{m}$ grid resolution covering wildfire-prone areas in the U.S., based on established classifications including Anderson's 13 fuel models~\cite{anderson1982aids} and Scott and Burgan's 40 fuel models~\cite{andrews2018rothermel}, which categorize vegetation by height and flammability to define burnability characteristics. However, existing fuel classification systems rely on static, coarse-resolution data that cannot capture monthly or quarterly environmental changes critical for disaster response and fire behavior analysis~\cite{finney1998farsite, finney2006FLAMMAP}.

\subsection{Problem Formulation}
Existing methods that adopt 3D representations and semantic segmentation fall short when providing solutions for large-area vegetated outdoor environments. It is essential to combine outdated Digital Elevation Models (DEMs) with current posed imagery to generate timely 3D terrain and fuel models. Terrain models are frequently outdated due to natural environmental changes, including prescribed burns, vegetation growth, and habitat shifts. Existing surveying frequencies remain insufficient for critical applications such as disaster mitigation.

Landscape changes in wildfire-prone environments occur more gradually than urban transformations, but the temporal lag between actual terrain and fuel changes and available digital representations creates significant gaps in environmental monitoring capabilities. Meanwhile, the large areas impacted by wildfire require methods suitable for sparse and spatially imbalanced imagery. Existing infrastructure often provides minimal view overlap, and even when overlapping views exist, establishing reliable correspondences between images remains challenging due to repetitive vegetation patterns and sparse, distinctive features.


\section{Method}
\subsection{Overview}
We propose an image-based pipeline for large-scale wildfire-prone landscape mapping. The pipeline integrates depth estimation, 2D-3D alignment, and image-based segmentation, leveraging the inherent stability of natural terrain. This stability enables historical DEMs to serve as reliable geometric priors for terrain reconstruction using current imagery.
Our rasterization approach efficiently represents large-scale areas, compressing $10\,\mathrm{km} \times 10\,\mathrm{km}$ landscapes into manageable 2D arrays. The raster-based structure supports multi-resolution spatial analysis, enabling both coarse semantic fuel mapping (30m resolution) and fine-grained terrain mapping (1m updated DEMs).

We first examine existing spatial representations and image-based 3D reconstruction approaches, analyzing their limitations for wildfire landscapes. Learning-based and photogrammetry-based methods are most widely adopted among 3D reconstruction techniques. We then investigate how these methods can be extended and combined with landscape-focused representations to address the unique challenges of vegetated environment monitoring.



\subsubsection{Objectives}
Given outdated DEM and real-time images, our goal is to provide an update to the DEM and fuel maps. This will allow the 3D semantic map of a large-scale wildfire-prone area to be kept up-to-date.

\subsection{Parameter Analysis for 3D Reconstruction}
We perform a computational complexity analysis that compares 3D representations based on raster, point, and volumetric data for large-scale terrain reconstruction. Table~\ref{table:compcost3D} presents the breakdown of the computational costs on different spatial scales and reconstruction methods. For raster-based DEM representations, computational complexity scales as the number of pixels, points, or other spatial registration methods. If the analysis is constructed to a square area of land where the width and length of the area are the same. Meanwhile, the precision (e.g., 1m gap between each data point) of 3D representation remains the same. As the coverage area increases, point-based methods suffer from cubic growth of width in point density requirements to maintain surface fidelity, while raster-based approaches maintain quadratic scaling with respect to width.

Our analysis reveals that raster-based representations achieve superior computational efficiency for large-scale vegetated terrain, with performance gains becoming more pronounced for larger scales. For the $10\,\mathrm{km} \times 10\,\mathrm{km}$ wildland areas typical in disaster monitoring applications, raster-based methods demonstrate a reduction in computational requirements compared to equivalent point cloud representations while maintaining sufficient geometric accuracy for large-scale outdoor vegetated terrain. It is noted that the raster 3D representation cannot capture hollow and concave features of landscape (e.g., overhangs, caves). But those scenarios are less relevant to wildfire propagation. 


\subsection{Photogrammetry-based DEM-integrated Method}
\subsubsection{Raster-pixel alignment}
\begin{algorithm}[tb]
\caption{Pixel-based On-raster Ray Tracing}
\label{alg:ray marching}
\textbf{Input}: Outdated 3D Models, real-time posed images\\
\textbf{Parameter}: $DEM[][]$, cam info $\theta$,$\phi$, $Cam_{elev}$, $img[][]$\\
\textbf{Output}: Depth map $depth[][]$, img-DEM$[][]$

\begin{algorithmic}[1] 

\FOR{$u$, $v$ in $img[][]$}
\STATE 
         $\vec{ray} \leftarrow F(\theta,\phi, u, v)
         $\label{alge:rayinit}
    \STATE $x, y, z = X_{cam}, Y_{cam}, Cam_{elev}$\label{alg:locinit}  
    \WHILE{$DEM[x][y] < z$}\label{alge:intersectionray}
 
        \STATE $x = x + \Delta_{X}(\vec{ray})$
        \STATE $y = y + \Delta_{Y}(\vec{ray})$
        \STATE $z = z + \Delta_{elev}(\vec{ray})$
    \ENDWHILE
    \STATE $depth[u][v]=|(x, y)-(X_{cam}, Y_{cam})|$\label{alge:depth}
    \STATE img-DEM$[u][v] = (x, y)$
\ENDFOR
\STATE \textbf{return} $depth [][]$, img-DEM$[][]$
\end{algorithmic}
\end{algorithm}

We designed ray tracing techniques to perform 2D-3D alignment between DEM raster pixels and image pixels. In Algorithm~\ref{alg:ray marching}, we demonstrate a method to perform this pixel-to-pixel matching. Given camera poses, camera locations, camera field of view (FOV), and terrain DEM, the algorithm outputs the depth map of the image and the 2D image-DEM alignment. As a result, each entry of the image-resolution 2D array is associated with DEM coordinates. In a selected DEM area, coordinates are represented by the distance from the origin divided by the DEM precision (i.e., 1 m).

In line~\ref{alg:locinit}, the location initialization for ray traversal is established in this designated coordinate system. The initial elevation is either collected from the camera's associated barometer or based on the DEM and common camera relative height offset (e.g., tripod height). In line~\ref{alge:rayinit}, the camera ray vector is created based on camera poses, camera FOV, and image resolution. Given the image pixel location, the ray is generated based on the proportional FOV offset. The ray direction defines the path along which the testing location traverses through space. The variables $(x, y)$ define the testing location used to query the terrain surface elevation, while 
elevation defines the ray elevation at the current testing location. In line~\ref{alge:intersectionray}, the stopping condition is defined as when the ray elevation falls below the terrain surface elevation. Thus, we identify the raster coordinate with which the image pixel aligns. We then calculate the depth of the pixel as the distance between the camera location and the found raster coordinate in line~\ref{alge:depth}.

\subsubsection{Vegetated perturbation topographical simulator}
We present a novel topographical landscape simulator for vegetated terrains with vegetation perturbation and burned landscapes. Our framework combines high-resolution digital elevation models with procedurally generated vegetation using Unreal Engine, enabling photorealistic rendering while providing precise ground truth data for geometry, semantics, and temporal evolution. We preserve geospatial consistency through real-world DEM data adapted to game engine input heightmaps, facilitating meaningful sim-to-real benchmarking.

Existing large-scale datasets and simulators, including KITTI~\cite{geiger2013vision}, 7-Scenes~\cite{glocker2013real}, CARLA~\cite{dosovitskiy2017carla}, and AirSim~\cite{shah2017airsim}, are predominantly designed for urban environments. While datasets for vegetated landscapes exist, they are typically aerial-based and lack the terrestrial-level imagery and topographical realism essential for evaluating 3D reconstruction or vegetation segmentation in large-scale outdoor environments. To address the lack of controlled vegetated terrain in remote wildland areas, we develop a custom simulation environment for outdoor vegetated scenes.


\subsubsection{Vegetation-shift images segmentation}
Building upon fuel type concepts~\cite{anderson1982aids}, we focus on three primary vegetation categories critical for wildfire assessment: grass (herbaceous), shrub, and tree (timber). By incorporating semantic segmentation into our image-based 3D mapping framework, we provide both geometric reconstruction and semantic understanding of vegetated landscapes, enabling more comprehensive environmental monitoring than geometric reconstruction alone.  Open vocabulary-based semantic segmentation methods (i.e., LSeg~\cite{li2022languagedriven}) for vegetation segmentation is executed by choosing the primary vegetation categories. Then, we utilized the segmentation results to create a fuel map.

\subsubsection{End-to-end 3D semantic fuel map}
After aligning DEM raster cell with corresponding vegetation on LANDFIRE-scale fuel maps, each pixel represents vegetation within a $30\,\mathrm{m} \times 30\,\mathrm{m}$ grid cell. While this contrasts with the 1 m precision of DEM maps, fuel maps are typically overlaid with DEMs during wildfire behavior analysis. The 2D fuel map can be projected into 3D space using DEM-derived elevation information. Assume $E(x, y)$ is the elevation map (e.g., from DEM), in meters. $L(x, y)$ is the fuel map, categorical. Coordinate $(x, y) \in \mathbb{Z}^2$ are pixel indices. $s = 30\,\mathrm{m}$ is the fuel map spatial resolution. We project each coordinate to 3D space using the elevation map and associate the corresponding fuel map cell label: $
P(x, y) = \left( x \cdot s,\; y \cdot s,\; E(x, y) \right), \quad L(x, y)
$

Fuels are categorized into three primary types: grass, shrub, and tree. Additional image categories include sky and mountain, which represent two major non-fuel domains present in wildfire-prone landscape imagery. To resolve multiple image pixels projecting onto the same fuel map cell (each representing a $30\,\mathrm{m} \times 30\,\mathrm{m}$ area), we use majority voting over the label space in Equation~\ref{equ:voting}.

\begin{equation}
    \hat{l}_{i,j} = \arg\max_{l \in \mathcal{L}} \sum_{ \mathrm{i-D}[u][v] \in \mathcal{N}(i,j)} \mathbf{1}_{L[u,v] = l}
    \label{equ:voting}
\end{equation}

$\hat{l}_{i,j}$ is the aggregated label assigned to fuel cell $(i,j)$, $\mathcal{L}$ is the set of possible fuel labels, $\mathcal{N}(i,j)$ is the set of $30\,\mathrm{m} \times 30\,\mathrm{m}$ area centered at $(i,j)$. i-D$[u][v]$ is the image-to-DEM alignment img-DEM$[][]$ we acquire from Algorithms~\ref{alg:ray marching}.

\subsubsection{Depth map with terrain priors}
Depth maps provide a efficient 3D representation where each pixel encodes the distance from camera to surface. We generate depth maps using Algorithm~\ref{alg:ray marching}, enabling direct 3D projection when combined with known camera poses.
Following FireLoc~\cite{fu2024fireloc}, we fuse neural depth estimation with DEM constraints through pixel-wise sampling and RANSAC regression. The approach can be enhanced using state-of-the-art monocular depth frameworks~\cite{piccinelli2025unidepthv2, bochkovskii2024depth}. This hybrid method leverages DEMs for stable terrain baselines while neural networks capture detailed surface variations, addressing training data limitations in outdoor environments. Performance degrades under significant occlusion conditions.

\section{Experiments}

\begin{figure*}[!t]

\begin{subfigure}[t]{.24\linewidth}
    \centering
\includegraphics[width=0.7\linewidth]{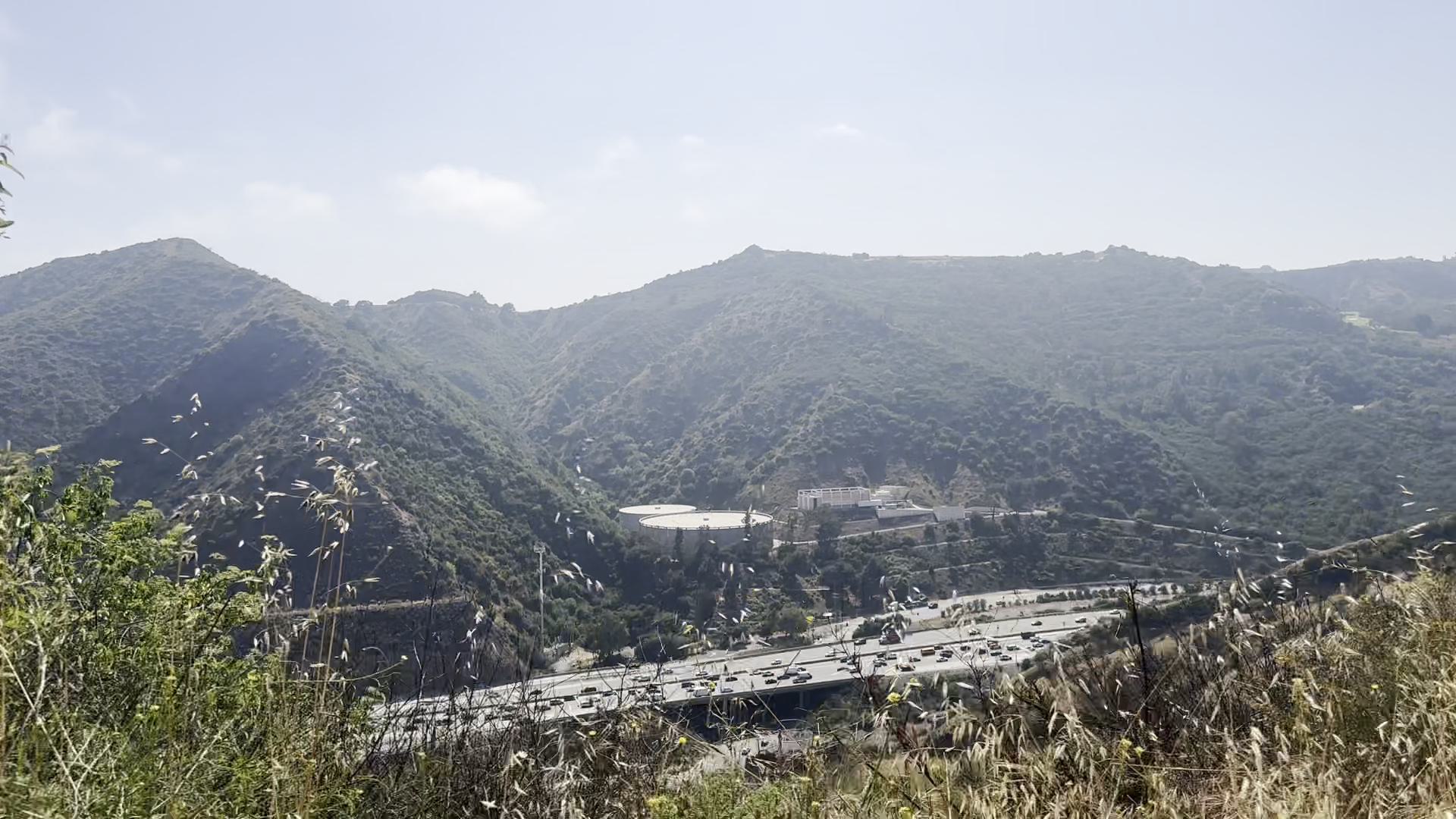}

\end{subfigure}
\begin{subfigure}[t]{.24\linewidth}
    \centering
    \includegraphics[width=1.1\linewidth]{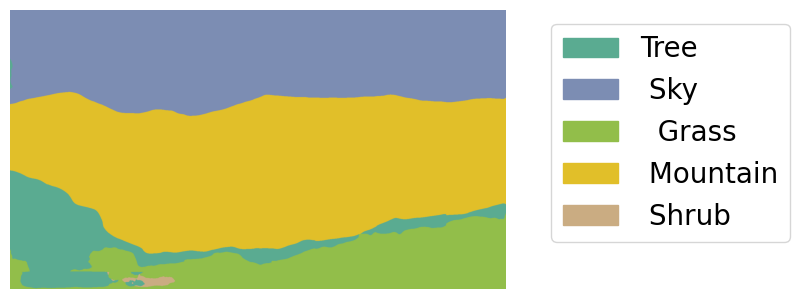}
\end{subfigure}
\begin{subfigure}[t]{.24\linewidth}
    \centering
\includegraphics[width=0.7\linewidth]{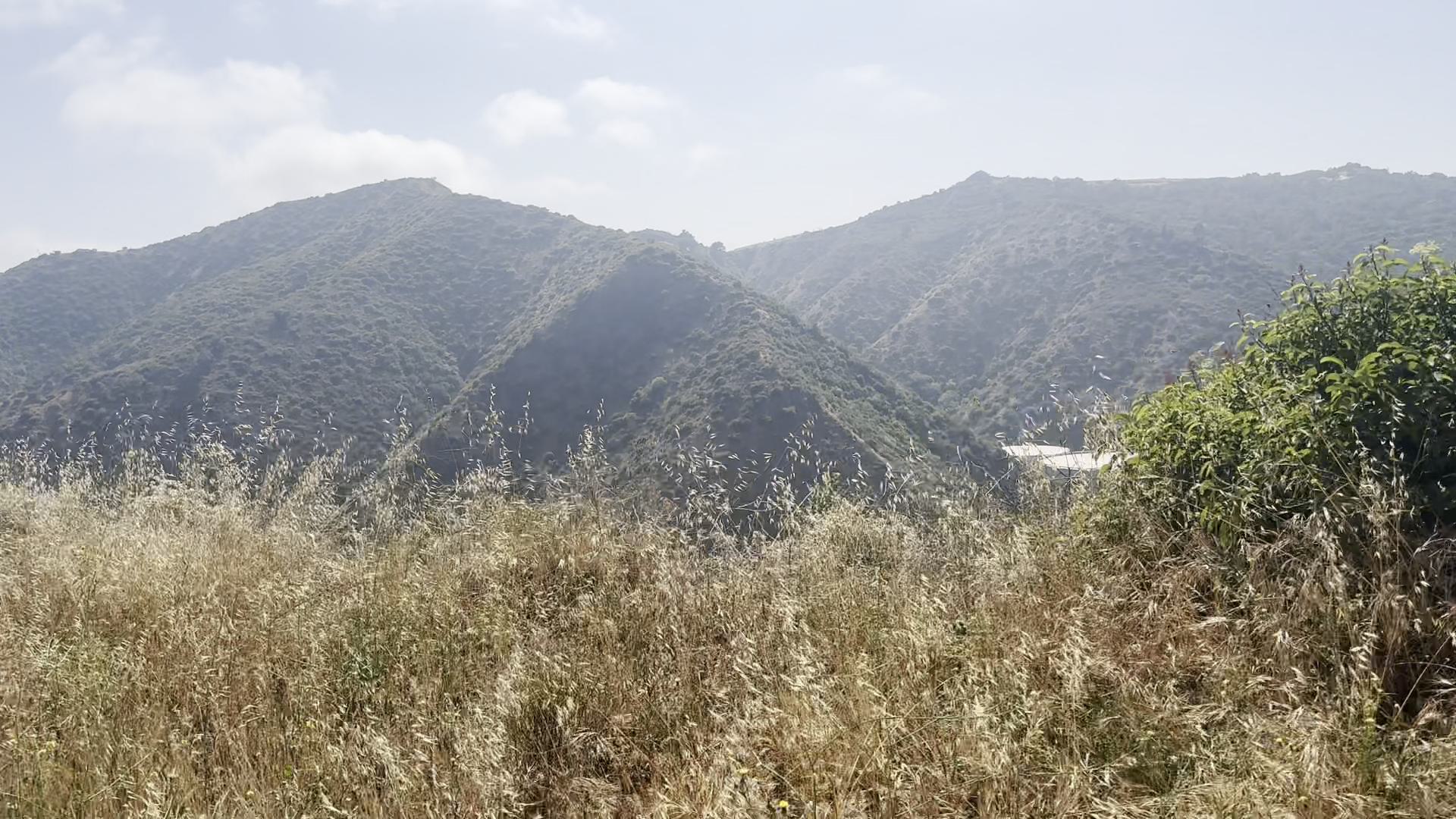}
\end{subfigure}
\begin{subfigure}[t]{.24\linewidth}
    \centering
    \includegraphics[width=1.1\linewidth]{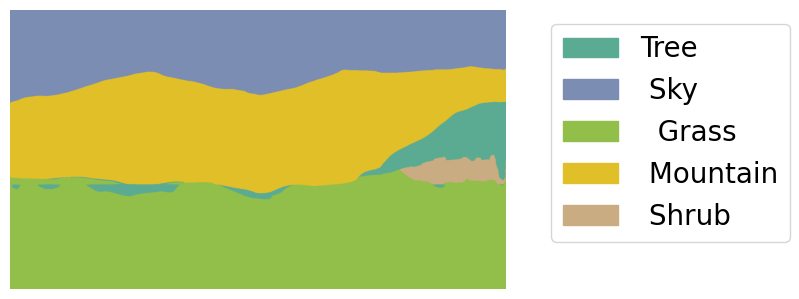}
\end{subfigure}

\begin{subfigure}[t]{.24\linewidth}
    \centering
\includegraphics[width=.7\linewidth]{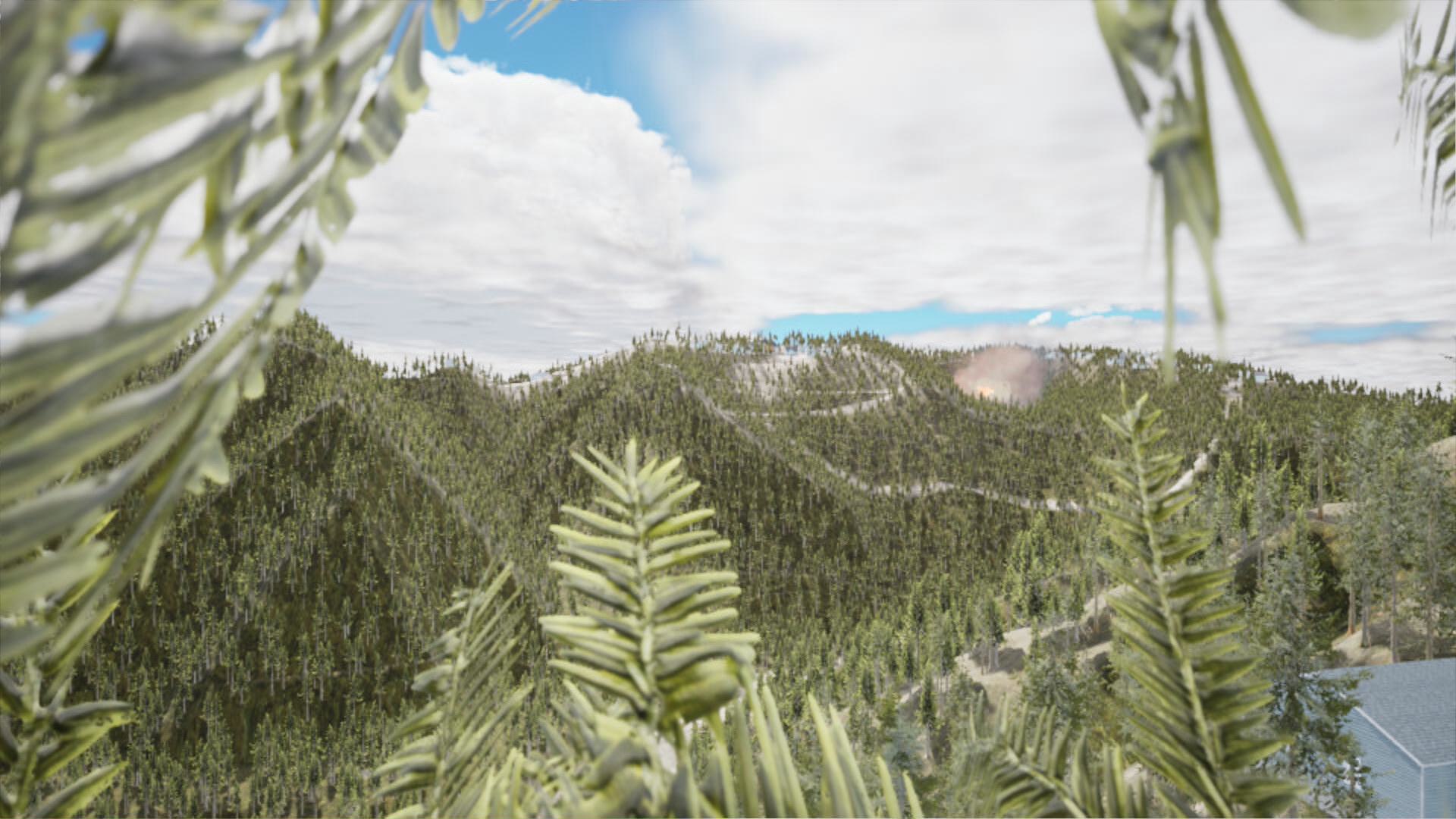}
\subcaption{Images}
\label{subfig:segscene1}
\end{subfigure}
\begin{subfigure}[t]{.24\linewidth}
    \includegraphics[width=1.1\linewidth]
    {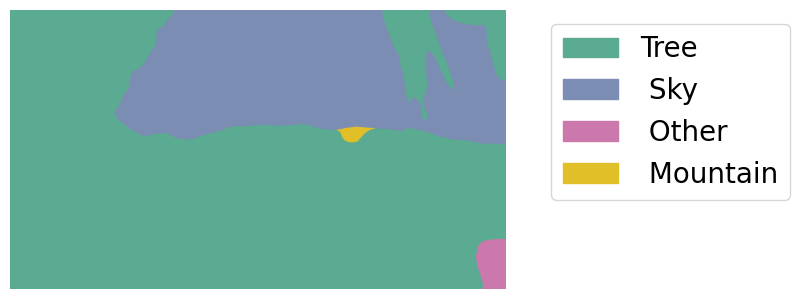}
\subcaption{ Segmentation}
\end{subfigure}
\begin{subfigure}[t]{.24\linewidth}
    \centering
\includegraphics[width=.7\linewidth]{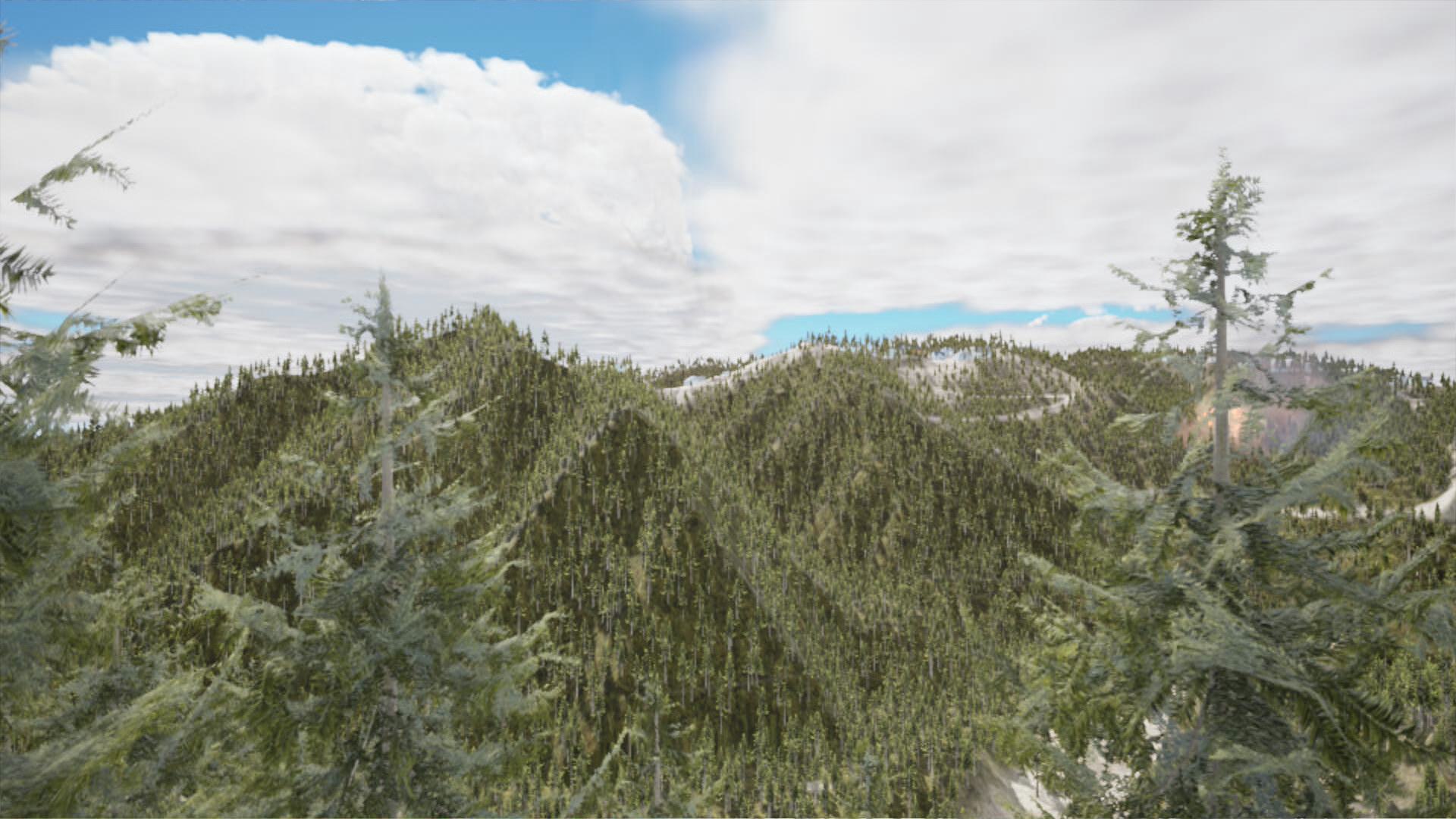}
\subcaption{Images}
\label{subfigure:segscene2}
\end{subfigure}
\begin{subfigure}[t]{.24\linewidth}
    \centering
    \includegraphics[width=1.1\linewidth]{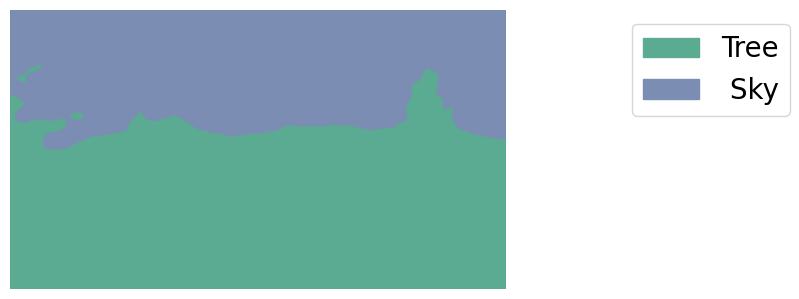}
    \subcaption{ Segmentation}
\end{subfigure}

\caption{We present 2 pairs of real-world, simulated images, and their respective image-based segmentation results. The segmentation are vegetation types based on wildfire fuel models. }
\label{fig:imageseg}
\end{figure*}

\subsection{Datasets}

\subsubsection{Site selection and terrain model}
We selected the Getty Fire site, focusing on the critical area spanning both sides of Interstate 405. Camera positions were established east of the highway with westward orientation to capture the region where the Getty Fire threatened to breach containment and cause extensive damage. Digital Elevation Model (DEM) data were obtained from OpenTopography.

\subsubsection{Simulated environment for large-scale vegetated scenes}

Our simulator generates realistic wildfire-prone landscapes with controllable vegetation parameters, enabling systematic evaluation of semantic mapping under varying environmental conditions. Built using Unreal Engine, the framework combines high-resolution digital elevation models with procedurally generated vegetation to produce photorealistic rendering while providing precise ground truth for geometry, semantics, and temporal changes.
The simulator replicates the same topographical area as our smartphone-collected real-world data, maintaining identical camera configurations. This enables direct sim-to-real comparison for vegetation change scenarios, including plant growth, prescribed burns, and other environmental modifications that alter landscapes over time. We evaluate both simulator performance and compare between simulated and real environments.

\subsubsection{Real-world scenes collection and alignment}
We collected real-world imagery from wildland locations using an iPhone 14 Pro to validate our simulation environment. The alignment between simulated and real scenes enables environmental parameter calibration, while the simulator provides ground truth geometry unavailable in real deployments. This hybrid approach balances realism with quantitative evaluation requirements.
Following established practices in location-based computer vision~\cite{kendall2015posenet}, we use smartphone sensors to capture images with pose and location data. Due to inherent drift in smartphone GPS, IMU, and barometer sensors, we manually filter results through visual inspection to ensure data quality. The simulator then generates corresponding images using identical pose and location information from the smartphone data.


\subsection{Experiment Details}
Simulations were conducted on NVIDIA RTX 3090 GPUs, while depth estimation experiments used dual NVIDIA A6000 GPUs. The simulation environment enables systematic perturbation of landscape parameters, vegetation density, terrain consistency, and cloud coverage condition.

\subsection{Evaluation Metrics}

\subsubsection{Sim-to-real accuracy}
We evaluate 3D reconstruction quality through image-level accuracy metrics, as terrain reconstruction ultimately depends on accurate depth estimation from individual viewpoints. Primary metrics include the Structural Similarity Index Measure (SSIM)\cite{wang2004image} for geometric consistency and the Learned Perceptual Image Patch Similarity (LPIPS)\cite{zhang2018unreasonable} for the evaluation of perceptual quality.
\subsubsection{Depth estimation and semantic segmentation}
The Depth estimation error can be calculated through errors, RMSE, meaning the distance between the actual location and the predicted location. While the semantic segmentation can be evaluated through the F-1 score or general fuel map cell classification errors. 

\begin{figure}[!t]
\centering
\begin{subfigure}[t]{.49\linewidth}
    \centering
\includegraphics[width=\linewidth]{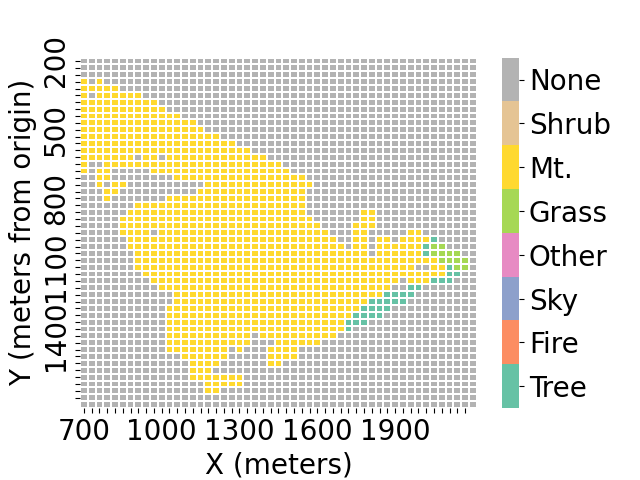}
\subcaption{Real-world vegetation}
\label{subfig:realsegmap1}
\end{subfigure}
\begin{subfigure}[t]{.49\linewidth}
    \centering
\includegraphics[width=\linewidth]{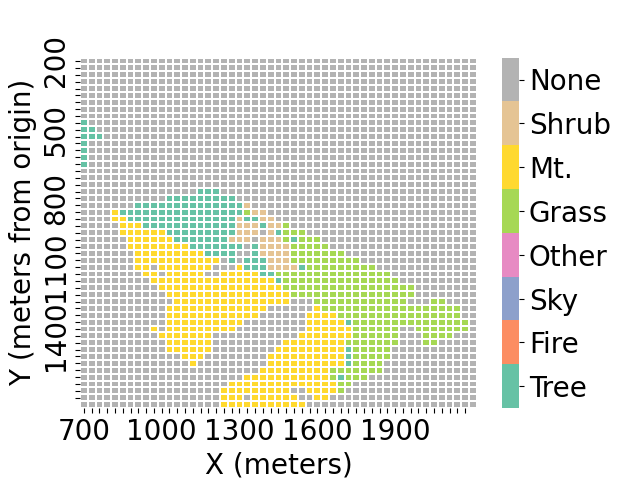}
\subcaption{Real-world vegetation}
\label{subfig:realsegmap2}
\end{subfigure}

\begin{subfigure}[t]{.49\linewidth}
    \centering
    \includegraphics[width=\linewidth]{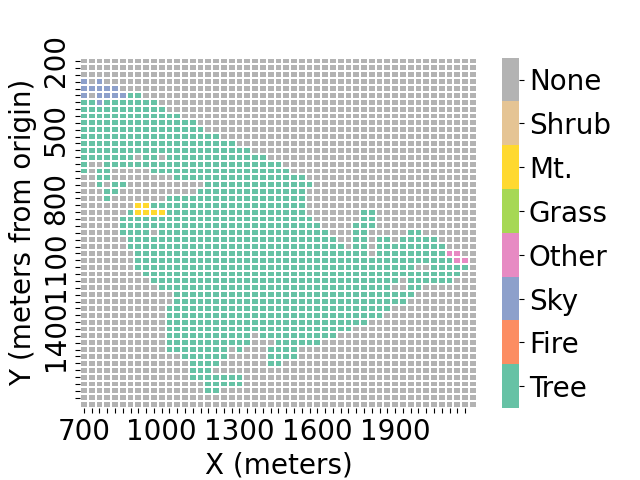}
    \subcaption{Simulated perturbation}
    \label{subfig:simsegmap1}
\end{subfigure}
\begin{subfigure}[t]{.49\linewidth}
    \centering
    \includegraphics[width=\linewidth]{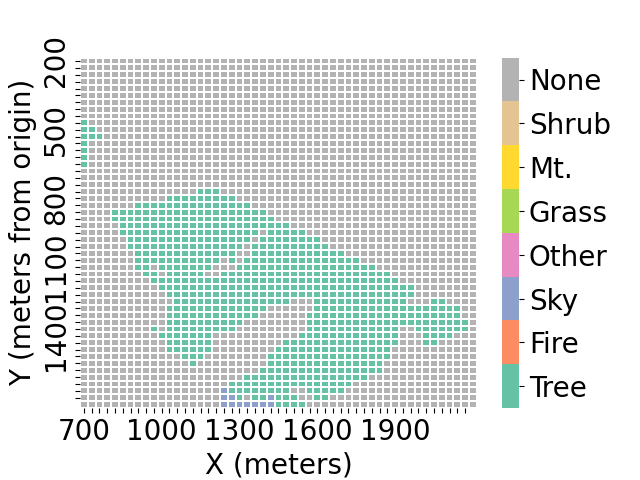}
    \subcaption{Simulated perturbation}
    \label{subfig:simsegmap2}
\end{subfigure}

\caption{Fuel maps comparing real-world and simulated camera setups over a $1500\,\mathrm{m} \times 1500\,\mathrm{m}$ area. Simulated data features vegetation perturbations.}
\label{fig:fuelmap}
\end{figure}

\subsection{Experimental Results}

\begin{figure*}[!t]
\centering
\begin{subfigure}[t]{.245\linewidth}
    \centering
    \includegraphics[width=.75\linewidth]{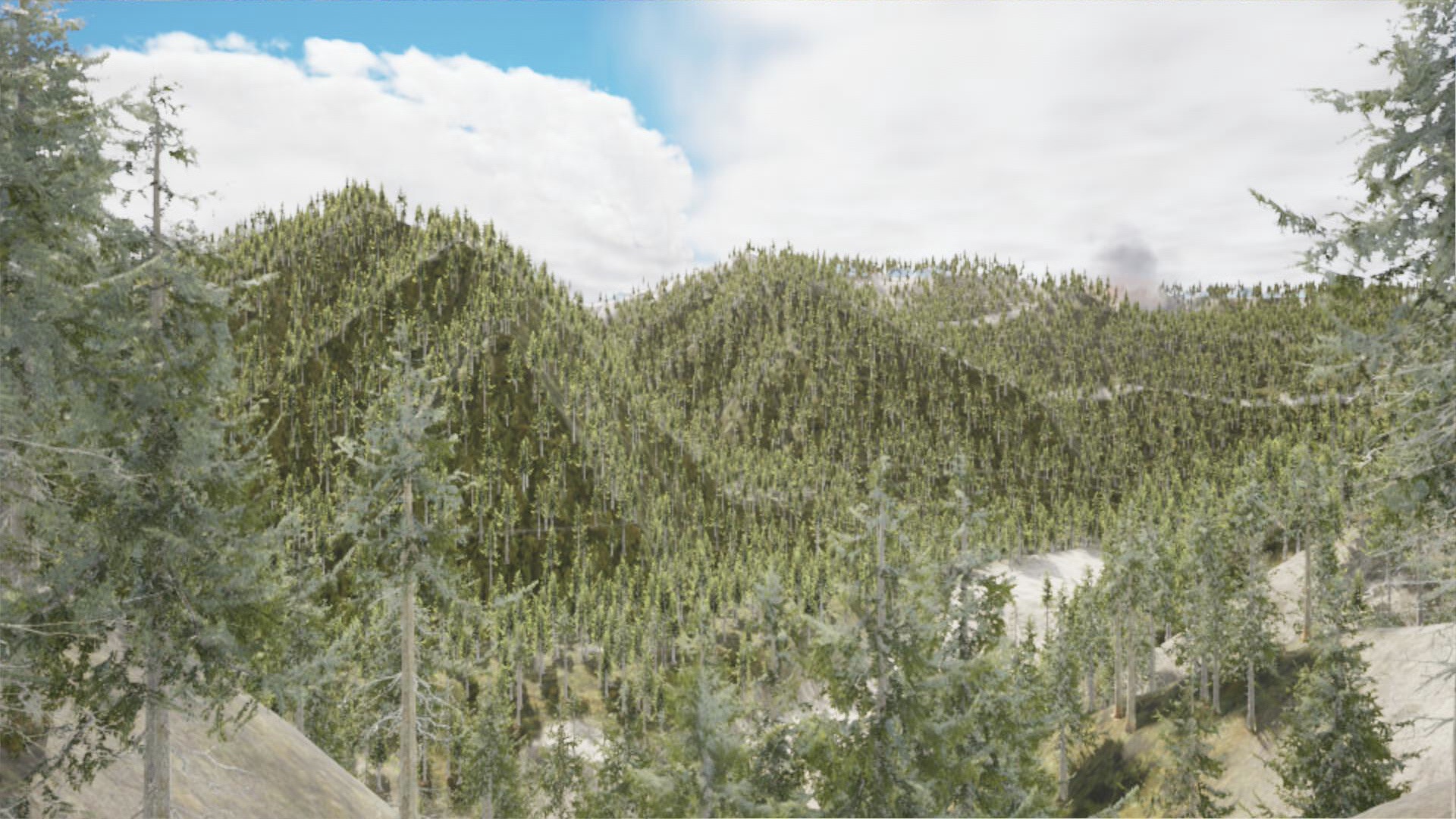}
\end{subfigure}
\begin{subfigure}[t]{.245\linewidth}
    \centering
    \includegraphics[width=1.1\linewidth]{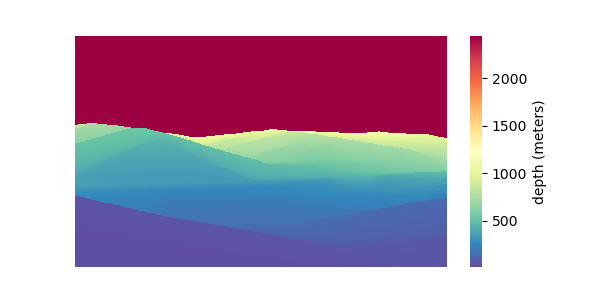}
\end{subfigure}
\begin{subfigure}[t]{.245\linewidth}
    \centering
    \includegraphics[width=1.1\linewidth]{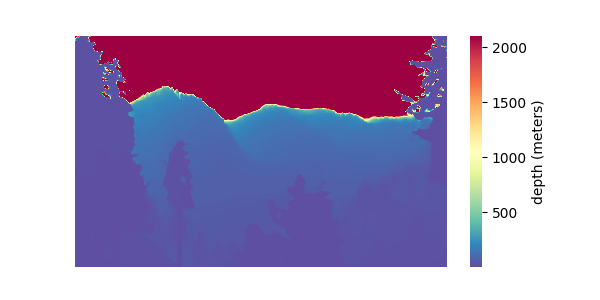}
\end{subfigure}
\begin{subfigure}[t]{.245\linewidth}
    \centering
    \includegraphics[width=1.1\linewidth]{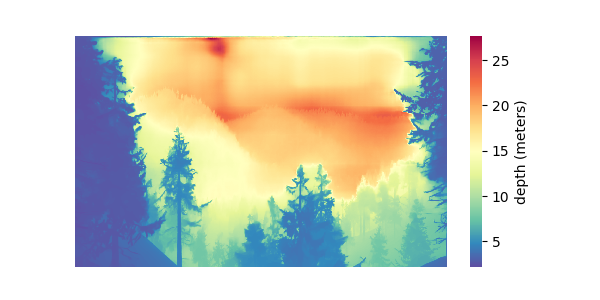}
\end{subfigure}



\begin{subfigure}[t]{.245\linewidth}
    \centering
    \includegraphics[width=.75\linewidth]{figures/sim2real/real00.jpg}
    \subcaption{Image}
\end{subfigure}
\begin{subfigure}[t]{.245\linewidth}
    \centering
    \includegraphics[width=1.1\linewidth]{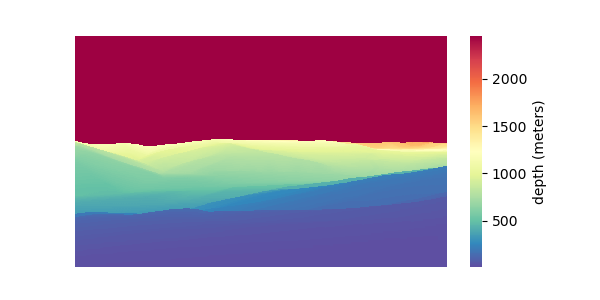}
    \subcaption{TopoDepth (ours)}
\end{subfigure}
\begin{subfigure}[t]{.245\linewidth}
    \centering
    \includegraphics[width=1.1\linewidth]{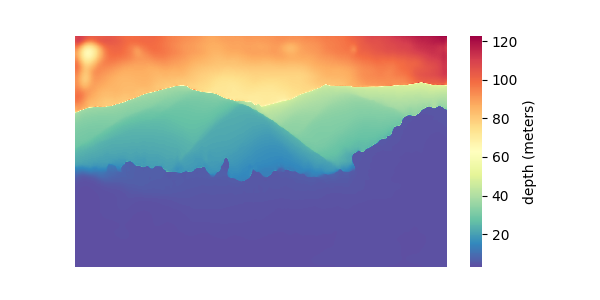}
    \subcaption{UniDepth}
\end{subfigure}
\begin{subfigure}[t]{.245\linewidth}
    \centering
    \includegraphics[width=1.1\linewidth]{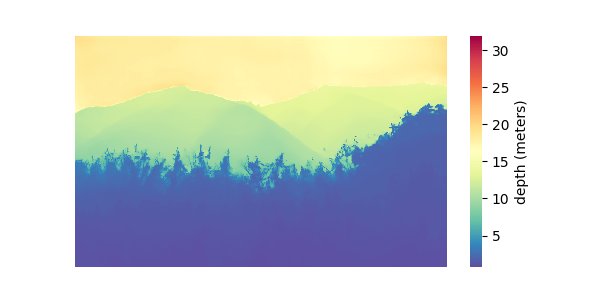}
    \subcaption{DepthPro}
\end{subfigure}

\caption{Monocular depth estimation in mountainous terrain. Our method outperforms UniDepth and DepthPro, which struggle with long-range scenes. While baselines capture fine details like vegetation outlines, they fail to estimate depth for mountain sections larger than 100 m away. TopoDepth provides accurate depth estimation for distant mountains. Results use non-aligned real-world and simulated image pairs. }
\label{fig:depthestimationcomp}
\end{figure*}
\subsubsection{Perception of the sim-to-real}
We present both simulated images and corresponding real-world captures to support the vegetation drift analysis. In Figure~\ref{subfig:segscene1} and Figure~\ref{subfigure:segscene2}, two representative scenes are shown. Each pairing is a real-world image collected with a calibrated camera and a simulated rendering generated using the same camera poses and geolocation data. The simulated results accurately capture key terrain features, including mountain ridges, ravines, and general landscape morphology. In addition, the simulator successfully reflects vegetation transitions, such as the shift from shrubland and grass to coniferous forest.

\subsubsection{Vegetation segmentation}
Image-based vegetation segmentation performs well in identifying the plant changes from the real-world shrubs to the simulated conifer terrain in the exact location. This is significant that the segmentation models can differentiate the shrub-like vegetation and conifer as presented in Figure~\ref{fig:imageseg}.  The camera locations are at grid coordinates (2218, 1101) for Figure~\ref{subfig:segscene1}, (2216, 1349) for Figure~\ref{subfigure:segscene2} in meters from the origin. The camera origin can be observed in Figure~\ref{fig:fuelmap}, where the 3D semantic segmentation results are presented.

\subsubsection{Fuel map}
Figure~\ref{fig:fuelmap} demonstrates successful fuel type classification, capturing the transition from conifer forest to shrubland. We evaluate the model in generic scenarios where vegetation transitions occur uniformly across the landscape (e.g., shrub/grass to conifer forest), and the model effectively identifies these broad vegetation shifts.

The resulting fuel classifications are projected onto 3D semantic maps using our proposed methodology. Camera coverage remains stable throughout data collection, as evidenced by the correspondence between real and segmented imagery: Figure~\ref{subfig:realsegmap1} (original shrub and grass), Figure~\ref{subfig:simsegmap1} (simulated coniferous forest), and Figure~\ref{subfig:segscene1} represent one viewpoint, while Figure~\ref{subfig:realsegmap2} (original shrub and grass), Figure~\ref{subfig:simsegmap2} (simulated coniferous forest), and Figure~\ref{subfigure:segscene2} represent another. This demonstrates the complete end-to-end pipeline success from initial image acquisition through fuel segmentation to final 3D fuel mapping.

\subsubsection{Monocular depth estimation}

We evaluate TopoDepth against state-of-the-art one-shot models, as shown in Figure~\ref{fig:depthestimationcomp}. UniDepth demonstrates general distance inference but lacks cross-scene consistency, particularly failing to maintain scale coherence across varying viewpoints. DepthPro differentiates near and far objects but cannot capture large-scale topographic structures, incorrectly classifying sky regions and distant peaks as mid-range depths. While prior methods achieve reasonable performance in controlled synthetic datasets, they exhibit severe degradation on real mountainous terrain where scale ambiguity and vegetation occlusion pose significant challenges.

TopoDepth produces spatially consistent depth estimates that accurately follow complex terrain transitions across varying elevations, maintaining errors within tens of meters—critical for wildfire-prone landscape mapping. Baseline methods are omitted from quantitative comparison due to depth predictions exhibiting 1-2 orders of magnitude greater error, miscalculating mountain ranges at 1000 meters as tens of meters away, rendering them unsuitable for deployment. By integrating DEM priors, our method overcomes the fundamental scale ambiguity in monocular depth estimation, achieving robust 3D understanding across simulated, real-world aligned, and image-raster aligned scenarios.

\subsubsection{Sim-to-real evaluation}
\begin{table}[t!]
\centering
\begin{tabular}{c|ccc}
\hline
                 & LPIPS  & SSIM & FID \\ \hline
Sim-to-real & 0.4437 &   0.3614   &   354.3967  \\
\hline
\end{tabular}
\caption{Sim-to-real evaluation.}
\label{table:sim2real}
\end{table}
We validate our FireLoc simulator against real-world fire imagery in a limited set of images (Table~\ref{table:sim2real}). As the first simulation to incorporate dynamic vegetation in wildfire modeling, our quantitative metrics demonstrate strong sim-to-real fidelity, establishing reliable benchmarks for fire prediction applications.

\section{Conclusion}
We present a novel multi-modal 3D semantic mapping method for large-scale outdoor vegetated environments. Our approach combines raster-based 3D representations with temporal information integration to achieve efficient terrain mapping in wildfire-prone landscapes. By adapting reconstruction techniques to address sparse features, temporal dynamics, and extreme scale requirements in vegetated environments, we demonstrate significant improvements in computational efficiency and accuracy. 
This work advances computer vision tools for environmental monitoring and disaster mitigation. 

\bibliography{aaai2026}

\end{document}